\documentclass[conference]{IEEEtran}
\IEEEoverridecommandlockouts

\usepackage{cite}
\usepackage{amsmath,amssymb,amsfonts}
\usepackage{algorithmic}
\usepackage{graphicx}
\usepackage{textcomp}
\usepackage{xcolor}
\usepackage{booktabs}
\usepackage{xurl}
\def\BibTeX{{\rm B\kern-.05em{\sc i\kern-.025em b}\kern-.08em
    T\kern-.1667em\lower.7ex\hbox{E}\kern-.125emX}}

\usepackage{eso-pic}
\newcommand\copyrightnotice{%
  \AddToShipoutPictureBG*{%
    \AtPageLowerLeft{%
      \raisebox{1em}{%
        \parbox{\paperwidth}{\centering\footnotesize\vspace{0.5em}
        This work has been submitted to the IEEE for possible publication. Copyright may be transferred\\
        without notice, after which this version may no longer be accessible.}}}}}
\makeatletter
\newcommand{\linebreakand}{%
  \end{@IEEEauthorhalign}
  \hfill\mbox{}\par
  \mbox{}\hfill\begin{@IEEEauthorhalign}
}
\makeatother

\begin{document}

\title{Integrity Detection and Characterization of Malicious Injections in RAVEN II}

\author{\IEEEauthorblockN{1\textsuperscript{st}Xingli Zhang}
\IEEEauthorblockA{\textit{School of Computing and Informatics} \\
\textit{University of Louisiana at Lafayette}\\
Lafayette, USA \\
xingi.zhang@louisana.edu}
\and
\IEEEauthorblockN{2\textsuperscript{nd} Diba Afroze}
\IEEEauthorblockA{\textit{School of Computing and Informatics} \\
\textit{University of Louisiana at Lafayette}\\
Lafayette, USA \\
diba.afroze1@louisiana.edu}
\linebreakand
\IEEEauthorblockN{3\textsuperscript{rd} Fei Hu}
\IEEEauthorblockA{\textit{Department of Electrical and Computer Engineering} \\
\textit{University of Alabama}\\
Tuscaloosa, USA \\
fei@eng.ua.edu}
\and
\IEEEauthorblockN{4\textsuperscript{th} Xiali Hei}
\IEEEauthorblockA{\textit{School of Computing and Informatics} \\
\textit{University of Louisiana at Lafayette}\\
Lafayette, USA \\
xiali.hei@louisiana.edu}
 }

\maketitle

\copyrightnotice

\begin{abstract}

The increasing adoption of robotic systems in surgery, together with the expanding range of procedures they can support and the growing level of autonomy they provide, has substantially increased the complexity of surgical robots. As these systems integrate more sensors, controllers, communication interfaces, and model-driven control components, their attack surface continues to expand. A compromise of the integrity of a surgical robot can therefore cause unintended robot behavior and potentially threaten patient safety.

In this paper, we characterize the detection boundary of malicious injections on RAVEN II using a public dataset that pairs the platform's telemetry with external high-resolution encoder ground truth. We identify three injection points spanning the command and observation paths and evaluate three injection patterns with increasing temporal dispersion. To capture different detection behaviors, we perform detection at two timescales: the window scale and the session scale. Rather than reporting detection rates at an arbitrarily chosen threshold, we quantify, for each injection point and injection pattern, the smallest end-effector deviation that can be resolved while maintaining an alarm rate acceptable for surgical operation. Our results show that detectability is strongly influenced by how the injected deviation is distributed over time. An abrupt step can be detected at deviations well below the 1 mm clinical tolerance, whereas the same overall deviation spread across a window or a session can remain hidden from single-window statistics. 
The open source code can be found at http://github.com/RAVENIIROS/RAVENIIIntegrity.
\end{abstract}

\begin{IEEEkeywords}
Surgical robot, RAVEN II, integrity monitoring, robot security.
\end{IEEEkeywords}
\section{Introduction}
Robot-assisted minimally invasive surgery (RAMIS) has gained wide acceptance among hospitals and patients worldwide for its precision and smaller incision. 
Teleoperated robot today is the most widely used mode: the surgeon works at a console, watching the live stream captured from the operation field, and sends commands over a local network to the host that drives the robot. The automated operations in surgical robots can reduce procedure time and errors that arise from surgeon fatigue during prolonged procedures~\cite{watanabe2016human}. The development of robotics~\cite{liang2026vivo} and the application of embedded AI (artificial intelligence)~\cite{knudsen2024clinical, AISurgery, Nvidiarobot} across different physical domains accelerates autonomy in surgical robots, particularly for highly repetitive tasks such as suturing and debridement.  Autonomous execution closes the operation loop on the telemetry itself without human participation. The surgical robot transformer, SRT-H, independently completed an extended gallbladder removal on a lifelike patient at JHU~\cite{JHUlive}. A notable subset of autonomous robotic surgery has successfully transitioned into clinical applications, such as venipuncture, hair implantation, intestinal anastomosis, total knee replacement, cochlear implant, radiosurgery, and knot tying~\cite{rivero2024autonomous}.  

Accurate end-effector motion is critical to surgical safety, particularly for fully autonomous systems, where no surgeon is present to provide real-time visual supervision. This makes end-effector motion an important target for attackers. The existing attacks mainly compromise the end-effector motion of the teleoperated robots via the communication channels~\cite{bonaci2015experimental, bonaci2015make}. An attack targeting the control system was subsequently proposed, and motivated a model-driven method for detecting and mitigating such attacks. The method monitors the command stream at the DAC layer with a preemptive dynamic model, integrating a second-order model of the motors and positioning joints to predict the next state and alarming when the predicted one-step change exceeds a threshold~\cite{alemzadeh2016targeted}. A learned context-aware variant extends that line~\cite{yasar2020real}. While those methods can detect the anomaly before a command is written, they still read commands from the control system and a malicious injection or modification written to the return path is invisible to them, such as the feedback data of autonomous robots and an injected offset in sensor readings at the physical layer~\cite{wang2026ghosttac}. A second gap concerns how detection performance is quantified. Previous work report detection probability at a chosen threshold, our work reports the caught smallest deviation at an alarm rate a surgery could accept.    

\begin{figure*}[t]
  \centering
  \includegraphics[width=0.85\textwidth]{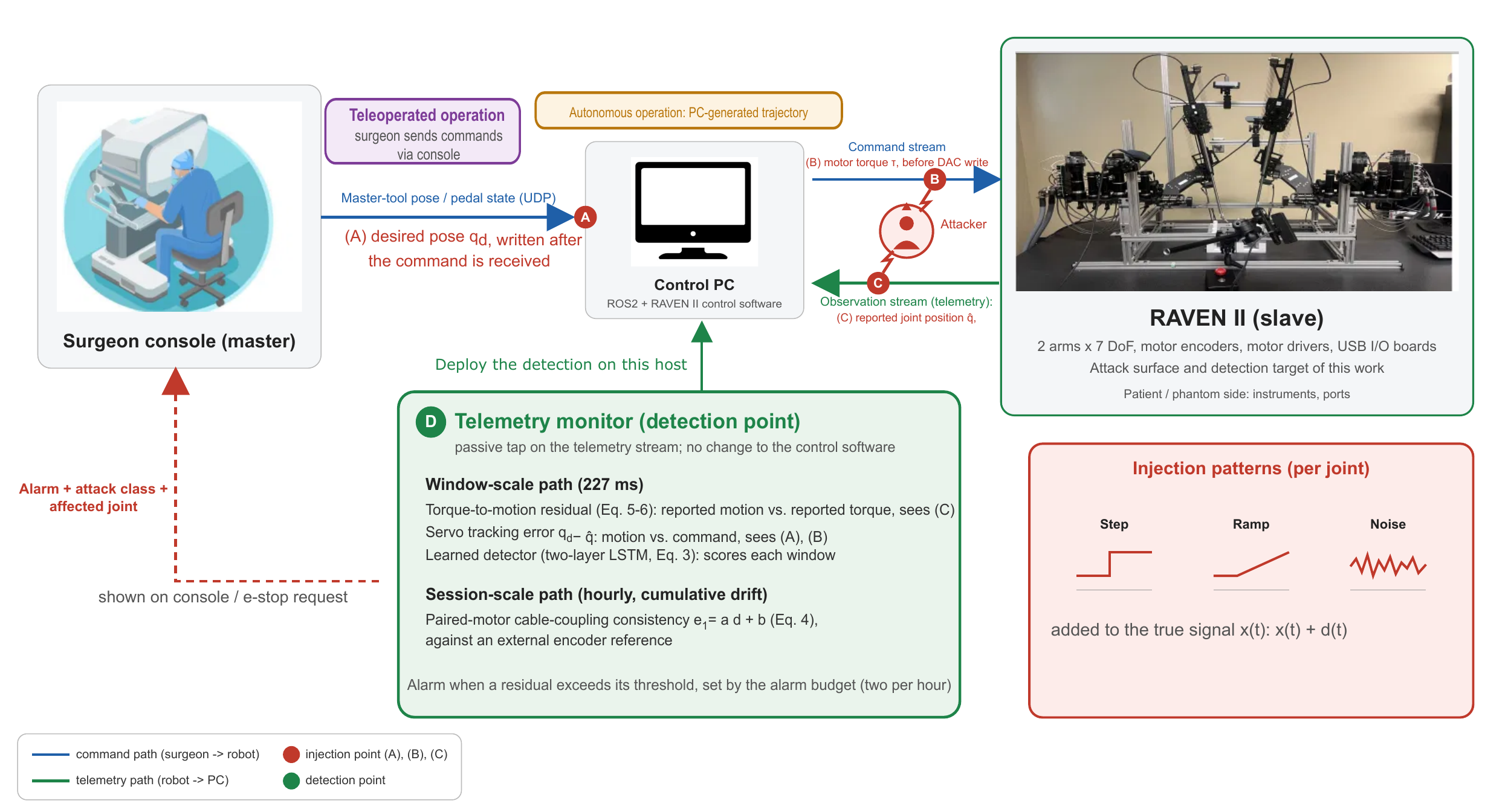}
  \caption{System overview. Surgeon commands flow from the master console to the control PC and on to the RAVEN II; ravenstate return parameter feedback of the robot to the surgeon (Teleoperation mode). Injection point              \textcircled{\small 1} perturbs the telemetry stream and injection point \textcircled{\small 2} perturbs the command stream (under tele-operated operation, on both the console--PC and PC--robot links; under autonomous operation, only on the PC--robot link). The telemetry monitor passively taps the same stream and raises an alarm when a residual exceeds a threshold set by a false-alarm budget.}
  \label{fig:overview}
\end{figure*}

To address these gaps, we investigate both the observability of deviations introduced along the return path and the minimum deviation that a telemetry-based monitor can resolve at an alarm rate acceptable for surgical operation. We use RAVEN II~\cite{hannaford2012raven}, an open surgical robot research platform operated in teleoperated mode, as a controlled testbed, while keeping fully autonomous surgical robots in view, where the controller acts on telemetry directly and no surgeon is present to catch what the monitor misses. Our evaluation draws on a public dataset of RAVEN II~\cite{peng2024efficient}\footnote{Dryad,  \url{https://datadryad.org/dataset/doi:10.5061/dryad.tqjq2bw84}}, collected on physical hardware, which uniquely pairs the platform's telemetry with external high-resolution encoder measurements.
We place injection points on both the command and observation paths. For each condition, we characterize the detection boundary as the smallest end-effector deviation that can be resolved at an alarm rate acceptable for surgical operation. We use 1 mm as the maximum clinically acceptable deviation, consistent with reported position-accuracy requirements for semi-autonomous abdominal surgery~\cite{peng2020real} and safety tolerances suggested for laparoscopic RAMIS~\cite{zhou2018new, tzemanaki2012towards}.

 Based on this attack-surface view and prior work, we identify three injection points where an attacker can modify the robot's data flow. (A) \textbf{Desired pose $(q\_d)$}, modified after the control software receives the operator's command and before inverse kinematics, determines the input to the kinematics, whereas (B) \textbf{Motor torque ($\tau$)}, modified after the safety checks and before the DAC write, determines the signal emitted by the servo; both points lie on the command path from the surgeon to the robot. (C) \textbf{Reported joint position ($\hat q$)}, modified on the return path from the end-effector to the control system, lies on the observation path. We evaluate these injection points under three temporal injection patterns and examine their detectability at both window and session timescales. Fig.~\ref{fig:overview} summarizes the overall system, attack surface, injection points, and detection setup.


We make three main contributions:

\begin{itemize}
    \item A characterization of the detection boundary on RAVEN II across three injection points, spanning the command and observation paths, and three temporal injection patterns, reporting the smallest end-effector deviation each detector resolves under both a threshold-free criterion and a fixed alarm budget. Command-path injections are realized through a validated servo model identified from the platform's own telemetry.

 \item Evidence that the temporal distribution of an injection has a substantial effect on detectability, and that the two reporting criteria can lead to different conclusions: a pattern that performs best under the threshold-free criterion can have no detection floor once an alarm budget is imposed. A sub-millimeter deviation detectable as a step can pass unnoticed when the same deviation is spread across a window.

\item A demonstration that no single feature or timescale covers all injection conditions. Each hand-designed residual sees only the injection points its channels compare, so command and observation paths require different features, and slow cable-related drift lies orders of magnitude below any single-window statistic, so it is resolved only by a session-scale consistency test against an external reference. The two paths are therefore complementary rather than complementary.

\end{itemize}

\section{Methodology}

\subsection{Datasets}

The dataset from Dryad has 19 channels: eight reported joint positions, eight motor torques, and first difference of the three positioning joints. Each recording is a 252-column log at 660\,Hz containing ground truth from external encoders (Avago AEDA-3300 at 80{,}000\,PPR on the rotational joints to measure the resulting cable degradation, Mercury II at 5\,\textmu m on the prismatic joint). We use two recordings: \texttt{record\_1\_different\_directions} on the injection experiments and \texttt{record\_3\_time\_decay} for the session-scale analysis. The latter includes three six-hour sessions under idle, unloaded, and 500 g loaded conditions. Since all the recordings are benign, we synthesize the injected data here. Injections are only applied to a single channel while other channels remain the original data. Three controlled tests confirm that the detector is reading the injection itself other than the synthetic data we add: (1) setting the injection magnitude to zero while retaining the attack labels; (2) restoring the attacked joint's columns to their clean values after injection; and (3) randomly shuffling the labels. The three tests yield chance-level performance across the three injection patterns, with AUCs of 0.503, 0.506, and 0.506, respectively.  

We work with a sliding observation window of 30 consecutive frames, and adjacent windows overlap by 25 of 30 frames. We sub-sample each window by five to 132 Hz and cut into the window with stride 5, but the splits of training data and evaluation data are by recording session to avoid the duplicate data in training set and test set.

\subsection{Injection patterns}
We define three injection patterns by temporal concentration: step, ramp, and noise, as shown in Eq.\eqref{eq:threepatterns}.

\begin{equation}
\label{eq:threepatterns}
\begin{aligned}
\delta_t &=
\begin{cases}
f(t), & t \geq t_0,\\
0, & t < t_0,
\end{cases}
\\
f(t) &=
\begin{cases}
m, & \text{Step},\\
s(t-t_0), & \text{Ramp},\\
\epsilon_t,\quad \epsilon_t \sim \mathcal{N}(0,\sigma_a^2), & \text{Noise}.
\end{cases}
\end{aligned}
\end{equation}

where $\delta_t$ denotes the joint angle offset in degrees, $m$ is the step magnitude, $s$ denotes the slope per frame, and $\sigma_a$ is the standard deviation of the injected noise. A step injects the whole desired deviation to a single frame and keeps it, appearing as one impulse. A ramp spreads the same deviation across every frame of the window, appearing as a continuous offset. Noise is independently sampled and added at each frame, yielding a zero mean, while only its variance encodes the attack. We draw \textit{$t_0$} from the middle third of the observation window, and report magnitudes as the window deviation for step and ramp and as RMS (Root Mean Square) for noise. These three therefore are all expressed in degrees of joint deviation and convert to millimeters at the end-effector through the forward kinematics~\cite{king2012kinematic}.  

The three patterns are defined in terms of joint position, which determines what an adversary can physically realize at each injection point. On the observation path, a step injection is straightforward: the adversary directly modifies the reported value, causing the joint position to jump within a single frame. 
On the command path, however, the situation is different. Although an adversary can issue a discontinuous command, as demonstrated in prior work~\cite{alemzadeh2016targeted}, the arm not deliver it. The auto-regressive with eXogenous~(ARX) model~\cite{ljung1998system} gives single joint step responses that reach 90\% of their final value within 6 to 17 frames at the 660~Hz rate, or 9 to 26~ms, which is one to four frames at the 132~Hz the detector reads. That rise is steep enough to be detected. A discontinuous command therefore appears in the reported position not as a step but as a short ramp, so the realizable position perturbations on the command path are better represented by the ramp pattern or a band-limited perturbation.

\subsection{Two Timescales}

We first identify how small an injection a window-scale detector can
resolve. The detector compares observed motion against predicted motion, so what limits it is the residual, the part of normal motion no model explains, rather than the size of the motion itself.
That leftover is measurable on the torque-to-motion relation, which is an explicit regression and therefore lets the split be read off directly. On the joint our step and noise
injections target, frame-by-frame displacement varies with a standard
deviation of $0.17^\circ$; the relation explains
$56\%$ of that variance and leaves a residual of
$0.11^\circ$. An injection has to move that residual by
appreciably more than $0.11^\circ$ before this feature separates it from normal motion.

This figure is the noise of the physics residual. Better
constructions can reach lower, and the learned detector resolves
$0.019^\circ$ on step injection. None reaches zero, because
what remains is physical: quantitation, mechanical backlash, and the
operator's own hand.


If an injection is slow enough, it is invisible for a window-scale detector, regardless of how much deviation eventually accumulates. For a window-level floor of $\ell$ degrees over $T_w$ seconds, any ramp whose rate satisfies 

\begin{equation}
r < \ell / T_w \quad \text{deg/s}
\label{eq:rate}
\end{equation}
is invisible. Taking $\ell$ to be the ramp floor of
Table~\ref{tab:patterns}, $0.174^\circ$, and with $T_w = 0.227$\,s and the deployment floor
of $0.174^\circ$ the threshold is
$0.766^\circ$/s, while natural cable degradation proceeds at
$3.7\times10^{-5}$ deg/s, four orders of magnitude slower, contributing
$8\times10^{-6}$ degrees per window.

Lowering the detection threshold does not solve the problem. To detect a change as small as \(8\times10^{-6}\) degrees per window, the threshold would have to be far below the normal variation, which is four orders of magnitude larger. The monitor would therefore trigger continuously even on a normal process. The fundamental limitation is the signal-to-noise ratio, and it applies to any statistic computed over a single window. 
We therefore introduce a slow path that provides a longer integration time. The two paths read different signals at different rates, run at different times, and an alarm from either is evidence on its own. The fast path scores each 227 ms window as it closes and triggers an alarm when that score crosses a threshold set from clean data at the chosen alarm budget. The slow path re-evaluates its calibration on the order of an hour, at a cost negligible beside the single window inference.

\paragraph{Fast path}
the window-scale path is a two-layer LSTM over one window, and its final hidden state feeds two heads. The classification head generates a score for each window. Every detection result in this paper is computed from that output in one of two ways: either by ranking windows relative to one another, measuring how often an injected window scores higher than a clean one without requiring a threshold, or by comparing each window’s score with a threshold derived from the clean-score distribution, thereby turning the score into an alarm decision. The regression head predicts end-effector displacement over the next 20 frames, but it is an auxiliary objective, computed on clean window only so that it learns undisturbed operations. Its weight exceeds that of the classification term, so the encoder’s representation is driven primarily by the need to predict future motion. We retain the classification term because training with classification alone may cause the model to rely on session-specific shortcuts that do not generalize across session splits. 
The input to the model includes 19 signals over 30 frames: eight reported joint positions, eight motor torques, and the first difference of the three positioning joints. 

\paragraph{Slow path} the session-scale path is two least-squares tests and validated on joint 1 only, but an adversary is free to write any joints. The linear relation construction can be applied to the other joints in principle. Adversary aims to modify the difference between the external encoder reading and the reported joint position by writing to that reported position. We instead read the two other motors. Analysis of the dataset revealed a pair of motors whose positional difference remains stable in motions. We attribute this stability to their shared cable path~\cite{hannaford2012raven}. Let \(d\) denote the difference between their reported positions. We observe that \(d\) responds to cable stretch without passing through
the reported position of joint~1, which is what an observation-path
adversary writes. We use $e_1$ for the reported position error on joint 1, the external encoder reading minus the value the platform reports. Cable stretch moves both $e_1$ and the pair difference \(d\). We fit 
\begin{equation}
e_1 = a\,d + b
\label{eq:slowcal}
\end{equation}
by least square and then freeze $a$ and $b$.  The consistency test monitors $e_1 - (a\,d + b)$ and raise an alarm when the error has moved without $d$ moving with it in every hour. A session runs six hours. To obtain $a$ and $b$, giving six points per session, we fit on the first three and evaluate on the rest. 

\subsection{Two consistency relations}
The learned model returns a score and indicate how small an injection is detectable but not which injection point a detector covers. A hand-designed relation answers that by construction, since what it can and cannot see follows from which channels it compares. 

The torque-to-motion residual indicates whether reported motion is consistent with reported force. 
\begin{equation}
\Delta \hat q_j \approx a_j \tau_j + b_j
\label{eq:f1}
\end{equation}
\begin{equation}
r_{j,t} = \Delta \hat q_{j,t} - (a_j \tau_{j,t} + b_j)
\label{eq:r}
\end{equation}

$\Delta\hat{q}_j$ is the frame-by-frame displacement reported by joint $j$, while $(\tau_j)$ is the motor torque of that joint. \(a_j\) and \(b_j\) are obtained by least-squares fitting on clean data and then frozen. $r$ is the residual, defined as the difference between the actual value and the model-predicted value. 

The servo tracking error $q_d - \hat q$, which indicates whether the motion is consistent with what the controller asked for.  $q_d$ denotes the desired joint position, while $\hat q$ denotes the reported joint position. Injection A modifies the value of $q_d$, injection B modifies the value of $\tau$, and injection C modifies the value of $\hat q$. 

This channel is not observable in the telemetry. At 132 Hz, \(q_d-\hat{q}\) does not exhibit the behavior of a tracking error: its standard deviation is comparable to the frame-by-frame displacement, its autocorrelation is 0.997, and its correlation with speed is negligible (\(|\rho|\leq0.16\)). With a sampling interval of 7.6 ms, the 6-ms phenomenon is not temporally resolved. At the native 660 Hz sampling rate, \(q_d\) leads \(\hat{q}\) by 4, 4, and 6 frames for the three positioning joints, respectively. Aligning the signals to compensate for this lead reduces the residual standard deviation by 54.5\%, 57.1\%, and 69.7\%, respectively. We therefore use the lead-aligned residual, with the lead estimated from the training sessions and then fixed for all subsequent evaluation.

\subsection{Command path simulation}
The position for each joint under malicious injections A is generated according to Eq. \eqref{eq:arx} from ARX model. 
\begin{equation}
\hat q[t] = \sum_{i=1}^{n_a} \alpha_i\, \hat q[t-i]
          + \sum_{j=0}^{n_b} \beta_j\, q_d[t-d-j]
\label{eq:arx}
\end{equation}
where $\alpha$ and $\beta$ are the coefficients fitted by least squares,
$d$ is the measured lag, and $n_a, n_b$ are orders chosen by held-out
validation. The resulting model achieves a held-out $(R^2)$ of 1.000000 with $(n_a=6)$ and $(n_b=4)$. However, this result alone is not sufficient to establish that the model captures the underlying system, since a model can achieve excellent prediction accuracy on data drawn from the same distribution without accurately representing the system dynamics. We therefore perform two additional checks that could have failed~\cite{ljung1998system}. First, a position servo should have a unit DC gain; otherwise, it would exhibit a steady-state offset. The identified DC gains are 1.012, 0.997, and 1.001. Second, the parametric frequency response should be consistent with an empirical transfer-function estimate that makes no assumptions about the model structure. At 1 Hz, the two estimates agree to three decimal places. The identified step responses reach 90\% of their final values within 12,
6, and 17 frames, the prismatic insertion joint slowest; that interval
bounds what an adversary can deliver as a position change on this path,
and it is two orders of magnitude shorter than a human correction loop. We simulate Injection A by adding the corresponding step response to the joint position. Injection B is modeled as a linear divergence based on the measured torque-to-displacement gain, assuming that the injected torque is not compensated by the controller. For Injection C, the injection onset is randomly selected within the middle of each window, and only each value in column $jpos$ is modified.

\section{Evaluation}
\subsection{Setup}
We train the fast path model on a DELL Precision 7680 with an RTX 4090, and measure detection time on RAVEN II's own control host, which has no GPU. 

\subsection{Evaluation parameters}

 We use area under curve (AUC) as the comparison criterion. The detector scores each window, with a higher score indicating a greater likelihood of an attack. Two windows are randomly selected: one injected and one clean. The AUC is the probability that the injected window receives a higher score than the clean window. 
  
 We use it because it makes floors comparable across features, patterns, and placements: a threshold-dependent criterion would confound each comparison with the choice of detection threshold.
 For each configuration we sweep injection magnitude over a logarithmic grid and evaluate the AUC on held-out sessions at each magnitude. We define the detection floor as the smallest magnitude at which the log-interpolated AUC reaches 0.9. AUC is not a deployment figure. It can remain high even when a small fraction of clean windows receive very high scores; once a target false-positive rate is specified, these high-scoring clean windows determine the detection threshold. A monitor evaluating 227 ms windows makes 15,850 decisions per procedure hour, so a per-decision false-positive rate has to be multiplied by that before it says anything about deployment.

\subsection{Necessity and validity of the session-scale path}
Over six hours under a 500 g load, the reported joint 1 accumulates a deviation of $0.8^\circ$ from the external encoder under benign operations. The fast path failed to detect the drift. We fit the physics model using the first hour of data and then froze its parameters. Across the subsequent six hours under load, we observed no significant change in the residual mean, standard deviation, or autocorrelation. It is because the drift signal, which is at $8\times10^{-6}$ degrees per window, is four orders of magnitude below the noise floor, as Eq.\eqref{eq:rate} predicts.

We also evaluate whether the relationship between $e_1$ and $d$ is physical or merely shared time trend. If they share only a trend, a fit made under one load will not be consistent with under another; if the cable drives both, it will. We therefore apply the frozen coefficients fit on the 500 g session to the unloaded one. The slope (a) only has a $6\%$ difference while the offset (b) moves by 0.121 degrees. This result indicates a shared physical cause: the slope reflects the physical coupling and transfers across the system, while the offset reflects the calibration. 

Therefore, two timescales are necessary. The session-scale detection covers the difference to 0.65 mm, where the reported joint position drifts while the channel sharing the same cable shows no corresponding change in cable length. It requires an external reference, such as an encoder. 

\subsection{Detection floors across injection points and streams}
Tracking error detects torque injection at 0.53 mm and desired-pose injection at 2.85 mm, while the torque-to-motion residual detects observation-path injection at 0.81 mm and, under execution-side torque logging, cannot see torque injection at all (Table~\ref{tab:coverage}). Torque injection is its best case, since a corrupted torque drives the arm away from its command and the error accumulates; desired-pose
injection is its worst, since the servo follows the corrupted command and leaves only a transient decaying within the settling time. A tighter servo therefore makes command injection more covert. No single feature spans the three points, and the two are complementary. 

A monitor's visibility depends on the stream it reads. Step is detected at 0.0389◦, or 0.29 mm at the tip, which is 2.05× coarser than on the observation stream but still inside the millimeter range. Ramp and noise are not detected at any magnitude we tried, as summarized in Table~\ref{tab:cmdstream}. The command stream is constrained by the motion a human operator can generate, whose bandwidth is limited to a few hertz. An instantaneous step therefore violates this temporal prior and is readily detectable. In contrast, a slow drift can resemble normal postural adjustment, while additional high-frequency content can be masked by tremor already present in the same frequency band. The observation stream has a different prior: the servo low-pass filters the operator's tremor before it reaches $\hat{q}$.

On the observation path, the learned detector resolves a step at 0.019° (0.14 mm at the tip), a ramp at 0.174° (1.30 mm), and noise at 0.0041° RMS (0.031 mm) under the threshold-free criterion (Table~\ref{tab:patterns}).

\begin{table}[t]
\centering
\caption{Each feature detects different injection points, with detection floors reported in millimeters of tip deviation. Neither feature covers all three points alone; together, they cover all three.}
\label{tab:coverage}
\scriptsize
\begin{tabular}{lccc}
\toprule
 & (A) $q_d$ & (B) $\tau$ & (C) $\hat q$ \\
\midrule
Arm physically moves          & yes    & yes       & no \\
$\Delta q \sim \tau$          & yes    & likely    & broken \\
\midrule
Torque-to-motion residual     & no     & likely no & $0.81$ \\
Servo tracking error          & $2.85$ & $\mathbf{0.53}$ & $1.35$ \\
\bottomrule
\end{tabular}
\end{table}

\begin{table}[t]
\centering
\caption{Detection floors on the command stream against the observation
stream, in degrees of joint deviation. The two streams are strong on
disjoint patterns.}
\label{tab:cmdstream}
\scriptsize
\begin{tabular}{lccc}
\toprule
 & command stream & observation stream & ratio \\
\midrule
step  & $0.0389$ & $\mathbf{0.0190}$ & $2.05\times$ \\
ramp  & none     & $\mathbf{0.1740}$ & --- \\
noise & none     & $\mathbf{0.0041}$ & --- \\
\bottomrule
\end{tabular}
\end{table}

\begin{table}[t]
\centering
\caption{The three injection patterns as applied to the observation path, with the floors they reach. All three edit the reported joint positions and leave every other channel as recorded, which is what makes them observation-path attacks and what lets the torque-to-motion residual see them. Under a threshold-free criterion noise has the best floor of the three; under any alarm budget our held-out data can estimate, none of the three has an operating point.}
\label{tab:patterns}
\scriptsize
\begin{tabular}{lccc}
\toprule
 & \textbf{step} & \textbf{ramp} & \textbf{noise} \\
\midrule
\multicolumn{4}{l}{\emph{Construction}} \\
Channels edited
  & $\hat q$ only
  & $\hat q$ only
  & $\hat q$ only \\
Joints targeted
  & $1$
  & $1$ or $2$
  & $1$ \\
Magnitude reported as
  & $|\delta_T|$
  & $|\delta_T|$
  & RMS \\
\midrule
\multicolumn{4}{l}{\emph{Floor at AUC $\geq 0.9$}} \\
Joint deviation
  & $\mathbf{0.0190^\circ}$
  & $0.1740^\circ$
  & $\mathbf{0.0041^\circ}$ \\
Tip displacement
  & $\mathbf{0.142}$\,mm
  & $1.300$\,mm
  & $\mathbf{0.031}$\,mm \\

\midrule
\multicolumn{4}{l}{\emph{Floor at a fixed alarm budget}} \\
Tip displacement  & none & none & none \\
\bottomrule
\end{tabular}
\end{table}

\subsection{Computational cost}
Inference costs  $0.662$,ms on a single thread of RAVEN-II's control host, matching the configuration expected for a deployed monitor. This leaves approximately one third of the $1$,ms control period unused by inference, while evaluating one non-overlapping window per control cycle occupies only $0.3\%$ of the available computation time. These measurements include only the forward pass; a deployed ROS node would additionally incur callback, serialization, and other runtime overheads that we did not measure.

\section{Future work \& Limitation}
\subsection{Predictive Detection}

The detector comprising two paths only reports the integrity of telemetry process after the motion. A predictive detection mechanism can be realized from what we have: the regression head forecasts displacement over the next 20 frames from clean windows, and its residual against realized motion is a detection statistic we have not evaluated. Forward motion is predictable and the ceiling is determined by the attack mode: everything in the telemetry is normal before the attacker modifies it. Thus, a detector operating on telemetry gains only the interval between the first corrupted frame and the point at which the accumulated deviation exceeds its threshold, estimated at three to ten frames for gradual injections and zero for a step. The significance of this interval depends on the control mode. Under teleoperation, tens of milliseconds are shorter than human reaction time, so the detector can support an automatic interlock but provides little additional protection; the surgeon can still visually identify the effects of the two command-path injections. Under autonomous execution, the same interval spans tens of control periods, with no visual supervision available as a fallback. A controller consuming corrupted telemetry can issue corrective commands, turning the injected telemetry error into physical motion. The case for a predictive detector is therefore strongest precisely where our evaluation is weakest: our dataset consists of teleoperated executions.

\subsection{Evaluation Limitations}
Two limitations bound the results. First, the floors for injections A and B depend on the identified servo model, because the trajectory that the arm would have followed under a corrupted command is not present in the recordings and must be reconstructed. The observation-path floor for injection C does not share this dependence, since it edits recorded values directly.

Second, an attacker with knowledge of the system can evade detection by the slow path, since the consistency test provides no protection against an adversary who manipulates both channels. By setting $\Delta d=\Delta e_1/a$, the adversary can leave the residual unchanged by construction. Across injections ranging from $0.05^\circ$ to $0.80^\circ$, the statistic remains at $0.145^\circ$, its no-injection value. When sweeping the compensation fidelity $\rho$, the attribution AUC falls below 0.9 at $\rho=0.90$, indicating that an estimate of $a$ accurate to within 10\% is sufficient to evade detection. For comparison, $a$ itself varies by 6\% between the two load conditions considered here. Thus, the required precision is lower than the quantity's own natural variation and can be obtained from a public dataset.

\section{Conclusion}
We characterized the integrity-detection boundary for malicious injections on RAVEN II by varying injection point, temporal pattern, and magnitude. Detectability depends on both where an injection enters the control loop and how its deviation is distributed in time. Under a threshold-free criterion the observation-path detector resolves
step injections to $0.142$\,mm and noise to $0.031$\,mm; on the command path, servo tracking error is the feature that sees torque and desired-pose injections, and the torque-to-motion residual is the independent evidence for the observation path. Detection must therefore run at two timescales. Window-scale statistics capture abrupt inconsistencies but cannot resolve slow cable degradation, which accumulates over hours and is caught only by session-scale evidence, at 0.65 mm here. No single feature covers all three injection points, and no single timescale covers both transient and accumulating deviations.

This boundary is a limit, not a guarantee. An attacker who knows the cable-coupling relation can manipulate both channels and preserve the slow-path residual. Telemetry-based integrity monitoring is thus best framed as characterizing which deviations remain observable under a given alarm budget, rather than as a binary question of detectability. Future work should evaluate predictive detection before corrupted telemetry reaches the controller and validate these boundaries under autonomous execution, where no surgeon provides visual supervision.

\section{Acknowledgments}
The authors used Claude Opus 5 to assist with grammar correction, spelling checking, and improving the clarity of the text in this manuscript. This tool was used strictly for language refinement under human supervision. No original research content, technical ideas, data, or scientific conclusions were generated by the AI system. The authors reviewed and edited all outputs and take full accountability for the accuracy and integrity of the final text.


\bibliography{cps,link}

\end{document}